# Modular Customizable ROS-Based Framework for Rapid Development of Social Robots

Mahta Akhyani[1], Hadi Moradi[1,2]

[1]School of Electrical and Computer Engineering

University of Tehran

Tehran, Iran

[2]Adjunct research professor, Intelligent Systems Research Center, SKKU, South Korea

mahta.akhyani@ut.ac.ir, moradih@ut.ac.ir

*Abstract*— Developing socially competent robots requires tight integration of robotics, computer vision, speech processing, and web technologies. We present the Socially-interactive Robot Software platform (SROS), an open-source framework addressing this need through a modular layered architecture. SROS bridges the Robot Operating System (ROS) layer for mobility with web and Android interface layers using standard messaging and APIs. Specialized perceptual and interactive skills are implemented as ROS services for reusable deployment on any robot. This facilitates rapid prototyping of collaborative behaviors that synchronize perception with physical actuation. We experimentally validated core SROS technologies including computer vision, speech processing, and GPT2 autocomplete speech implemented as plug-and-play ROS services. Modularity is demonstrated through the successful integration of an additional ROS package, without changes to hardware or software platforms. The capabilities enabled confirm SROS's effectiveness in developing socially interactive robots through synchronized cross-domain interaction. Through demonstrations showing synchronized multimodal behaviors on an example platform, we illustrate how the SROS architectural approach addresses shortcomings of previous work by lowering barriers for researchers to advance the state-of-the-art in adaptive, collaborative customizable human-robot systems through novel applications integrating perceptual and social abilities.

## I. Introduction

Designing robotic systems capable of natural socio-emotional interaction poses significant cross-disciplinary challenges. While existing frameworks address individual aspects such as locomotion, sensing, or user interfaces, fully realizing the potential of human-robot collaboration necessitates a unified architectural approach integrating capabilities from robotics, computer vision, artificial intelligence, and interactive design [1-3].

While many social robots such as Furhat [4] and NAOqi [5] rely on proprietary software that limits customization, recent frameworks have aimed to facilitate more open-ended development. Platforms like Tiago [6] provide ROS integration for mobility but have more limited support for social capabilities. Additionally, prior approaches often treat robotics, artificial intelligence, and user interfaces as separate concerns rather than enabling tight integration across domains [7-8].

We present the Socially-interactive Robot Software platform (SROS), an integrated open-source framework addressing these shortcomings through its layered structure and modular encapsulation of specialized skills as ROS services. SROS's modular, layered structure seamlessly bridges robotics via ROS with web and mobile interfaces using standard messaging and APIs. By separating concerns across interconnected layers while facilitating tight synchronization of data streams, SROS enables the systematic design of collaborative robot behaviors and percepts. Specialized AI skills such as facial analysis, speech processing, and multi-modal fusion are implemented as plug-and-play ROS services. This flexibly deploys advanced competencies across diverse robotic morphologies. SROS thus lowers barriers to rapid prototyping while promoting the reuse of computationally intensive perceptual modules.

Through demonstrations, we aim to evaluate SROS's potential to streamline the development of assistive systems exhibiting natural, context-aware socio-emotional skills. Its holistic yet specialized architectural approach addresses cross-disciplinary integration challenges, empowering researchers to push the forefront of human-robot interaction. Through demonstrations showing synchronized multimodal behaviors on an example platform, we illustrate how the proposed framework aligns with interaction design best practices [9-10] through its adaptability to different embodiments and behaviors. SROS's architectural approach aims to address shortcomings of previous work by lowering barriers for researchers to advance the state-of-the-art in collaborative customizable systems through novel applications that integrate perceptual and social abilities.

## II. A. Prior work

Many social robots, such as Nao [11] and Pepper [12], have proprietary integrated software stacks that constrain the range of behaviors and use cases. Tiago uses a ROS-based modular robot software framework with more focus on mobility rather than social capabilities. Based on our search on the research in this area, there has been limited work on customizable ROS-based frameworks tailored for social robotics. However, existing approaches often treat robotics, artificial intelligence capabilities, and user-centered design as separate concerns. While existing robotics frameworks provide navigation, control, and perception tools, developing socially competent robots requires tight integration across disciplines. The use of standardized messaging and well-established tools also improves cross-platform compatibility and code reuse. This makes it well-suited to industrial and consumer robotics with browser or app-based interactions. While ROS is a widespread middleware for robotics applications, it does not inherently support integrated user interfaces or advanced perceptual skills. Web-based frameworks like Blockly [13] enable graphical robot programming but isolate user interfacing from robot-centric ROS capabilities. Platform-specific frameworks such as Furhat and NAOqi concentrate capabilities into singular robotic platforms without facilitating cross-platform or modular development.

Both SROS and the platform proposed by Elfaki et al. [14] aim to provide a framework for social robot development that integrates various technologies and domains. While their work focuses on a specific domain of edutainment for hospitalized children, SROS can support different types of social robot applications. Additionally, SROS relies on standard messaging and APIs to coordinate the communication between the layers, while their proposed framework uses a centralized server that might introduce latency and dependency issues. A list of similar frameworks is presented in Table 1.

TABLE 1: COMPARISON OF FRAMEWORKS OR DEVELOPING SOCIAL ROBOTS WITH THE PROPOSED FRAMEWORK

| Platform | Web service Integrated. | Android Integrated | ROS-Based | Explicity for Social Robots | Explicity for Social Robots | Supports Persian |
|---|---|---|---|---|---|---|
| SROS | ✓ | ✓ | ✓ | ✓ | ✓ | ✓ |
| Elfaki. et al. | ✓ | ✓ | ✓ | ✓ | ✓ | - |
| AMIRO [15] | ✓ | - | ✓ | ✓ | ✓ | - |
| Misty [16] | - | ✓ | ✓ | ✓ | ✓ | - |
| FURHAT [4] | - | - | ✓ | ✓ | ✓ | - |
| OPSORO [17] | - | - | ✓ | ✓ | ✓ | - |
| RobWork [18] | - | - | ✓ | - | ✓ | - |

Building upon existing work, SROS presents a unified architectural approach that systematically bridges the gap between these domains. Its layered structure and tight, standards-based integration of ROS and REST APIs enable the seamless development of socially interactive robot behaviors. This promotes a modular yet synchronized implementation of capabilities across various disciplines. Specialized perceptual and interactive robot skills can also be reused more readily through SROS's approach of encapsulating them as ROS services. This distinguishes SROS in addressing the need for an integrative yet flexible framework.

In this paper and study, we show the capability of integrating different social interaction capabilities using SROS by integrating Persian speech recognition and text-to-speech (TTS) capabilities, including an experimental Persian GPT2-based module. This feature makes SROS stand out as a robotics platform that enables research with populations not fluent in English by supporting a second language to a significant extent.

Through the demonstration of its capabilities, we show how SROS can accelerate the research and development of socially intelligent robots. Its integrated yet modular framework has the potential to spur innovation through rapid prototyping of next-generation human-robot collaboration.

## II. SYSTEM ARCHITECTURE

### A. Overview

We present a novel modular and customizable software architecture for developing social robot capabilities using ROS. The proposed design consists of key layers to enable the integration of advanced perceptual and reasoning skills required for natural HRI.

A distributed graphical user interface layer allows for personalized remote access to various devices. A centralized controller promotes contextual robot behaviors through asynchronous coordination of independent components. Computational pipelines are partitioned across ROS nodes to optimize processing.

Core robot functions like multimodal sensing, activity recognition, and emotional expression are abstracted as interoperable services. This standardized interface approach streamlines the rapid mixing of capabilities.

The architecture advocates distributed pipelines for synchronized real-time processing of streaming data. Modularity allows customized interface adaptations and behavioral modifications.

This flexible framework is uniquely suited to provisioning next-generation robot assistants, companions, and tutors with the high-level perceptual and socio-emotional sophistication needed for meaningful social engagement with humans.

The proposed architecture aims to advance the state of the art by empowering researchers to quickly prototype and rigorously evaluate novel applications of human-robot interaction. We believe its synergistic design choices can significantly accelerate progress in the field.

### B. System Design

The SROS architecture incorporates modular hardware and software components tailored for social robot capabilities. The overall system architecture consists of four main layers as shown in Fig. 9:

1. Graphical User Interface (GUI) Layer: Allows remote monitoring and control of the robot via a web interface.
2. Web Layer: Manages user requests, parses the GUI, coordinates services between GUI and other layers, and handles the database

3. ROS Layer: Handles low-level robot I/O, sensor processing, and actuator control, while providing reusable perception, analysis, and generation capabilities.
4. Hardware Layer: Handles the robot's hardware (Camera, mic, actuators, etc.)

In the following, each layer is explained in more detail.

*B.1 Graphical User Interface Layer:*

The GUI is a web-based control panel developed using HTML, CSS, and JavaScript. This control panel provides remote access to monitor and control the robot. It allows users to define temporary sequences of multimedia behaviors combining specific face and sound pairings. It connects to roscore through ROS Bridge and subscribes to needed topics on the page load. A live video feed from the robot's camera can be viewed through image_raw and image_raw/landmarked topics with options to turn facial landmark analysis graphics overlay on or off. Virtual joystick controls enable teleoperating the robot's movements by sending geometry_msgs/Twist messages on cmd_vel_wheel topic as shown in Fig. 1. Future work may include additional features to further enhance the remote control and visualization capabilities.

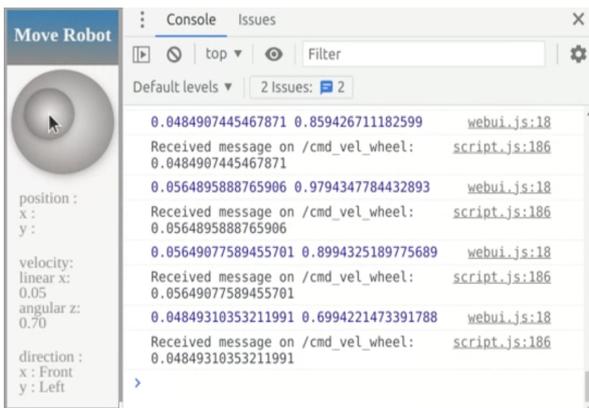

Figure 1: Commanding robot movement using the implemented virtual joystick on the interface via ROS Bridge communication.

*B.2. Web Layer:*

The layered web services architecture plays an integral role in realizing the systematic vision behind SROS. By providing a stable and standardized integration layer based on RESTful APIs and asynchronous messaging, it facilitates tight yet modular coordination across robotics, perception, and user interfaces.

The web layer itself consists of four different layers:

1. Database
2. Django Web application
3. Python WSGI
4. Web Server

The Django framework acts as the central coordinator, exposing APIs that different system elements can interact with independently. This treats each capability as an independently developed and tested module. Gunicorn and Nginx play crucial roles in facilitating the seamless integration of modular elements in real-time, enabling the robot to achieve holistic functioning. By leveraging Gunicorn and Nginx, the system can effectively handle incoming requests, route them to the appropriate components, and ensure optimal performance and scalability.

Most importantly, this architecture supports SROS's synchronized dataflow approach by treating perception, actuation, and interface elements as interconnected yet specialized domains. Standardized communication protocols promote flexible yet tight collaboration both within and across these domains.

In this way, the layered web services stack is the core to realizing SROS's overarching goal of systematically integrating robotics, AI, and HRI through a modularized yet unified methodology.

The database provides a standardized method for storing attributes related to system elements. It is currently used to store emotion behavior profiles, which link visual and audio demonstrations to affective states. These multimedia assets serve as examples for perceptual modules, showcasing a promising approach for future expansion by incorporating additional emotion or behavior sequences in the database for enhancing the robot's affective expression handling.

Robot definitions outlining available actuators and sensors are also stored. This enables front-end interfaces to dynamically generate interactive elements tailored for each robot morphology.

In the future, the database could expand its role by centrally defining profiles for individual skills. For instance, parameters for specific motors. This would treat low-level components as independent services that higher-level architectures assemble based on declared capabilities.

By providing a common interface for defining and relating attributes, the database complements the standardized communication enabled by the web services infrastructure. This treatment of data modeling supports the methodology's extensibility and facilitates optimizing complex robot behaviors through centralized access to specifications.

The database exemplifies techniques for systematically coordinating traditionally separate research areas like perception and control within a unified robotic framework.

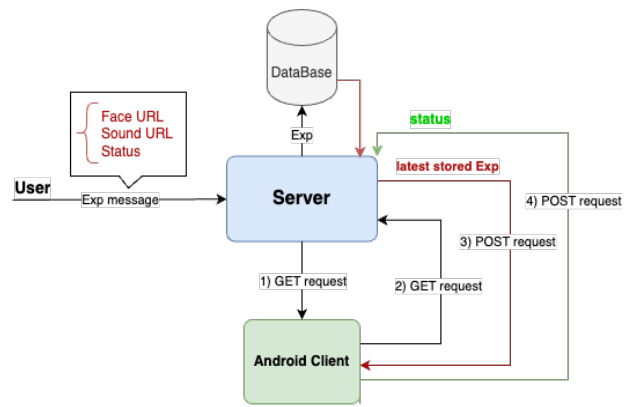

Figure 2: The Exp request handling and updating behavior pipeline (The requests are numbered in the sequence of occurrence.)

*B.3 ROS Layer:*

The ROS layer executables follow the following structure pattern:

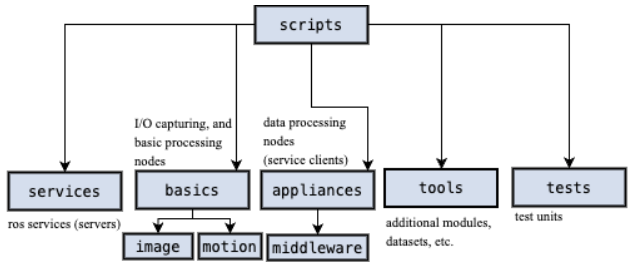

Figure 3: The ROS layer executables directory structure pattern

The system distributes computationally intensive tasks like computer vision and audio processing across separate ROS nodes, improving parallel efficiency. Nodes in the "basic" directory handle robot I/O inputs and initial processed data, that need to be published and accessed constantly for research purposes. The nodes in "middleware" manage to route heavier data processing pipelines. These nodes call the corresponding services in the "services" directory and publish the output to the defined topic. This design facilitates customization in two ways. First, users can redefine individual nodes to update algorithms or I/O. Second, launch file configuration determines node inclusion/exclusion in experiments.

Fig. 4 illustrates the proposed ROS-based computational graph for real-time multi-task computer vision processing. At the start of the pipeline, raw color images and the camera information are published by the video stream node from an RGB camera. The images and the camera information are published to the image_raw and camera_info topics respectively for downstream consumption.

The core image processing is handled by the OpenCV client node, which bridges ROS and OpenCV. It transforms subscribed image and calibration data into an OpenCV image format before publishing preprocessed images to the image_cv2 topic. Here we must emphasize that since ROS does not support multidimensional array messages, OpenCV data such as the frame and landmarks are first transformed from Numpy arrays to lists of lists, and then published as custom messages. Other system limitations are also overcome following this methodology.

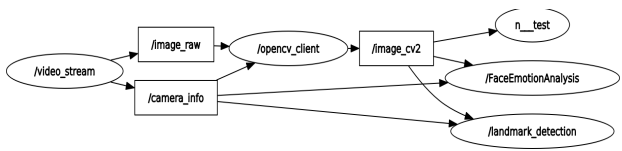

Figure 4: Image processing pipelines for synchronized real-time processing of streaming data. The video_stream node publishes the camera input on image_raw topic and the calibration info (e.g., image width, height, etc.) on the camera_info topic. The opencv_client node then subscribes to both topics and publishes the converted Image messages to 3D arrays on the image_cv2 topic. Both FaceEmotionAnalysis and landmark_detection nodes subscribe to this topic for further image processing.

Two computer vision pipelines branch off from the image_cv2 topic. The FaceEmotionAnalysis node implements the DeepFace module [19] to perform facial emotion classification. In parallel, the Landmark_detection node utilizes Media Pipe [20] for facial landmark localization, publishing full-body pose data accurately as shown in Fig. 6.

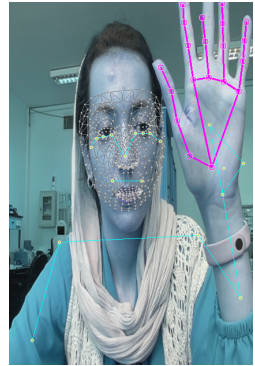

Figure 6: Demonstration of real-time landmark detection accuracy (rqt_image_view frame of the /image_raw/landmarked topic)

Detected landmarks then get utilized in other sensory data processing services such as gaze_detector service as shown in Fig. 7.

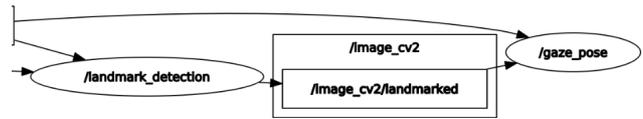

Figure 7: Utilizing the detected landmarks for gaze detection service

The gaze pose estimator service as a downstream node of our example pipeline (e.g. the computer vision pipeline), is responsible only for the gaze detection functionalities and calculations. Its service estimates the head's position (pitch and yaw values), and then projects the pupil location on the 3D plane of the face to estimate the gaze position. This methodology helps keep a wide variety of functionalities separate which helps significantly while expanding the architecture by adding new modules.

This modular architecture supports various extensions and applications. For example, face recognition can be added by calling relevant services or subscribing to the image_raw/landmarked topic[1] to bypass early processing steps. Our goal is to provide a configurable toolkit that equips researchers with fundamental algorithms and interfaces through ROS, rather than an all-in-one solution. This is intended to streamline the prototyping of social robotics concepts by furnishing the necessary building blocks to pace development. The extensible design aims to facilitate expedited workflow in collaborative robotics research.

The remaining capabilities of the proposed system follow the same data processing protocol of vision. The proposed ROS-based functionalities include broadcasting camera image converted to OpenCV image (as custom-defined messages for lists of lists since ROS doesn't yet support 3D arrays), broadcasting landmarked camera input image, broadcasting audio input as audio_common messages, Random Forest Classification for audio sentiment analysis, HAZM library [21] for transcribed speech word tokenizing, ParselMouth Praat speech feature extraction, Google speech recognition API for speech-to-text, Microsoft Edge TTS for text-to-speech, Google Media Pipe for full body landmark detection and then for gaze estimation, and DeepFace for facial expression estimation for expression imitation.

---

[1] The topic on which the landmarked image is being published as Sensor_msgs/Image messages. A sample image view of this topic is as shown in Fig. 7.

Finally, the test units for the whole system can be found in the tests directory. Each unit is written for a specific node using the standard unit test library.

*B.3 Hardware Layer:*

Hardware Layer: The modular architecture of SROS is designed to facilitate research on a wide range of hardware platforms without limitation. Commercial humanoid robots like Nao [22] can be used directly. Alternatively, developers are free to implement the platform on their own experimental robots in lab, home, or educational settings.

SROS even enables initial prototyping and testing of perceptual/interactive skills using only common hardware like a phone, without requiring a robotic platform. This lowers barriers to entry and brings human-robot interaction research within reach of most communities.

SROS, as a platform independent architecture, has implemented support for several common robot components out of the box. It incorporates serial communication functionality, allowing integration of Arduino-based features and sensors. Support is also included for Dynamixel servo motors through native ROS control capabilities. General purpose I/O access enables controlling basic actuators like DC motors through GPIO pins.

In the future, SROS aims to expand its dynamic robot capabilities with features like autonomous navigation. But the core modular vision is to remain hardware-agnostic, focused on advancing State-of-the-art HRI methods through flexible and accessible software tools. Overall, SROS strives to bring sociable robotics innovation full circle from research to real-world applications.

## III. KEY TECHNOLOGIES

SROS features a set of core perceptual and interactive capabilities implemented as reusable modules.

- Computer Vision: Landmark detection using Media Pipe extracts 2D body/face key points from camera images in real-time.

  DeepFace performs facial analysis to classify emotion expressions based on detected landmarks.

- Speech Processing: Audio feature extraction with Praat [23] calculates MFCCs and prosodic metrics from microphone input.

  Google Cloud Speech API [24] performs Persian speech recognition.

- Dialogue & NL Understanding: Natural language parsing and GPT-2 based response generation enables basic conversational skills in Persian.

These specialized capabilities are implemented modularly as independent ROS nodes for reusable integration across robot morphologies through SROS's standardized interfaces and messaging. Future works will target the continual expansion of the platform's social competence toolkit.

## IV. DEMONSTRATION

We experimentally validate SROS's capabilities using both an Nvidia Jetson Nano-powered social robot platform with RGB-D fish eye camera and microphone array deployed in our lab, as well as a virtual machine running Ubuntu 20.04 on a separate computer with an HD camera and laptop microphone array. Data streams were collected from these hardware configurations.

Landmark detection from video streams enables tracking of face/body poses. DeepFace accurately classifies facial expressions in real-time.

For computer vision tasks, SROS broadcasts camera images from an RGB-D camera. Our landmark detection node processes these using Media Pipe, publishing full-body pose data accurately as shown in Fig. 7.

Recorded audio samples are input to the audio processing pipeline. Features are extracted using Praat and emotions are classified via a Random Forest model. Results are published over ROS.

Speech recognition is demonstrated via test utterances processed through the Google STT API. Transcriptions are returned and output on the web UI.

User input of simulated emotions to the web interface triggers behaviors, such as playing matching audio clips through physical speakers.

The Android app receives video/audio streams over WiFi and outputs the status of the streaming action.

Gaze estimation from the integrated landmark and pose data is also simulated. The published outputs are as shown in Fig. 8. This illustrates SROS's capability to enable the development of socially competent robot behaviors.

```
^Cmahta@mahtavm:~/OSSRP/cleancoded$ rostopic echo /gaze_position/gaze_dir
data: "left up"
---
data: "left up"
---
data: "left up"
---
data: "left up with head being  up"
---
```

Figure 8: Demonstration of gaze pose estimation service

The modular design of SROS supports deployment on different robotic hardware, from simulated platforms to those with varied sensors and actuators.

Modularity is demonstrated through the successful integration of an additional ROS package, ROS4HRI [25], without changes to hardware or software platforms. This demonstrates the framework's flexibility and potential for rapid prototyping of socially interactive robots.

## V. CONCLUSION

In this work, we presented SROS, an extended framework built upon ROS, for developing socially intelligent robots through tight integration of robotics, AI, and interface design. SROS addresses the challenges of cross-disciplinary interactions through its layered yet synchronized architectural approach.

Key contributions include the use of ROS for robotics combined with REST APIs and topics to seamlessly bridge lower and higher levels and the extensive feature support for the Persian language. Implementing specialized perceptual skills as ROS services promotes modular reuse across morphologies.

Experimental validation demonstrated SROS's capabilities for computer vision, speech processing, and mobile accessibility. Its modular structure accommodated additional packages without platform changes.

Moving forward, ongoing work involves expanding the toolkit of social competencies like gesture recognition, emotion models, and tactile sensing. Applications targeting human-robot collaboration in assistive domains are also being explored.

## VI. Acknowledgement

The authors like to acknowledge the feedback and cooperation of different teams working on the social robots in the Advanced Robotics and Intelligent Systems lab. Specifically, Pegah Soleiman for helping to test the proposed SROS on RoboParrot. Also, Shahab Nikkhoo for helping to test SROS on Hooshang, a mobile humanoid robot. Finally, Bijan Mehralizadeh to test the system on the comprehensive autism screening system which include RobotParrot, the sensorized toy car, and the light wheel.

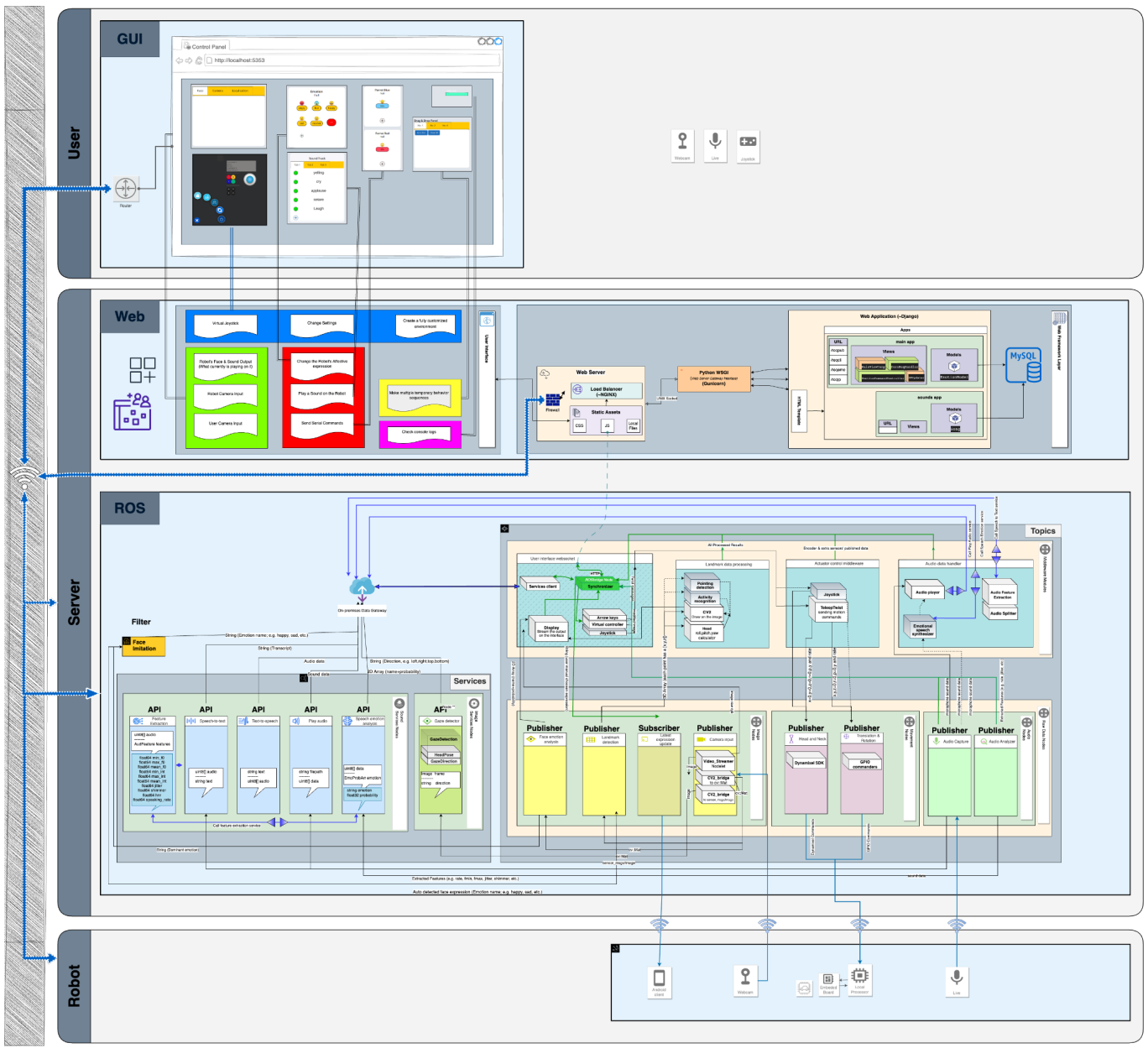

Figure 9: Overall architecture of the proposed framework